\documentclass[journal]{IEEEtran}

\usepackage{amsmath}
\usepackage{amsfonts}
\usepackage{tikz}
\usepackage{multirow}
\usepackage[mathscr]{eucal}
\usepackage{algpseudocode}
\usepackage{algorithm}
\usepackage{amssymb}

\tikzset{>=latex}

\DeclareSymbolFont{bbold}{U}{bbold}{m}{n}
\DeclareSymbolFontAlphabet{\mathbbold}{bbold}
\newcommand{\ind}{\mathbbold{1}}

\newcommand{\tail}{\mathrm{tail}}
\newcommand{\head}{\mathrm{head}}
\newcommand{\cost}{\mathrm{cost}}
\newcommand{\edit}{\mathrm{edit}}
\DeclareMathOperator*{\argmax}{\arg\!\max}
\DeclareMathOperator*{\argmin}{\arg\!\min}
\newcommand{\inedges}{\text{in}}
\newcommand{\outedges}{\text{out}}

\newcommand{\term}[1]{\textbf{#1}}

\newcommand{\thetaenc}{\Theta_\text{enc}}
\newcommand{\thetadec}{\Theta_\text{dec}}

\newcommand{\htcomment}[1]{\textcolor{magenta}{[#1 --HT]}}
\newcommand{\llcomment}[1]{\textcolor{blue}{[#1 --LL]}}
\newcommand{\klcomment}[1]{\textcolor{red}{[#1 --KL]}}
\newcommand{\kgcomment}[1]{\textcolor{green}{[#1 --KG]}}
\renewcommand{\htcomment}[1]{\ignorespaces}
\renewcommand{\llcomment}[1]{\ignorespaces}
\renewcommand{\klcomment}[1]{\ignorespaces}
\renewcommand{\kgcomment}[1]{\ignorespaces}

\begin{document}

\title{END-TO-END NEURAL SEGMENTAL MODELS \\ FOR SPEECH RECOGNITION}
\author{
    Hao Tang,~\IEEEmembership{Student Member, IEEE},
    Liang Lu,~\IEEEmembership{Member, IEEE},
    Lingpeng Kong,~
    Kevin Gimpel,
    Karen~Livescu,~\IEEEmembership{Member, IEEE},
    Chris Dyer,
    Noah A.\ Smith,
    and~Steve Renals,~\IEEEmembership{Fellow, IEEE}%
\thanks{Hao Tang, Liang Lu, Kevin Gimpel, and Karen Livescu are with
    Toyota Technological Institute at Chicago, Chicago, IL 60637 USA (email:
    \{haotang,llu,kgimpel,klivescu\}@ttic.edu).}
\thanks{Lingpeng Kong is with
    Carnegie Mellon University, Pittsburgh, PA 15213 USA (email:
    lingpengk@cs.cmu.edu).}
\thanks{Chris Dyer is with
    Google DeepMind, London, UK and
    Carnegie Mellon University, Pittsburgh, PA 15213 USA (email:
    cdyer@google.com).}
\thanks{Noah A.\ Smith is with
    University of Washington, Seattle, WA 98195 USA (email:
    nasmith@cs.washington.edu).}
\thanks{Steve Renals is with
    University of Edinburgh, Edinburgh, UK (email:
    s.renals@ed.ac.uk)}
}

\markboth{IEEE JOURNAL OF SELECTED TOPICS IN SIGNAL PROCESSING, VOL. X, NO. Y, AUGUST 2017
    }{TANG et al.: END-TO-END NEURAL SEGMENTAL MODELS FOR SPEECH RECOGNITION}

\maketitle

\begin{abstract}
Segmental models are an alternative to frame-based models for sequence prediction, where hypothesized path weights are based on entire segment scores rather than a single frame \klcomment{edited} at a time.
Neural segmental models are segmental models that use neural network-based weight functions.
Neural segmental models have achieved competitive results for speech recognition, and their end-to-end training has been explored in several studies.
In this work, we review neural segmental models, which can be viewed as
consisting of a neural network-based acoustic encoder and
a finite-state transducer
decoder.
We study end-to-end segmental models with different weight functions,
including ones based on frame-level neural classifiers
and on segmental recurrent neural networks.
We study how reducing the search space size
impacts performance under different weight functions.
We also compare several \klcomment{edited} loss functions for end-to-end training.
Finally, we explore
training approaches,
including multi-stage vs.~end-to-end training and multitask training that combines 
segmental and frame-level losses.
\end{abstract}

\begin{IEEEkeywords}
segmental models, connectionist temporal classification,
end-to-end training, multitask training
\end{IEEEkeywords}

\section{Introduction}


Automatic speech recognition (ASR) has been treated as a graph search problem
since its early development~\cite{J1976}, and
the graph search approach has been popularized by the use
of hidden Markov models (HMM)~\cite{rabiner1989tutorial,bahl1983}.
Given a sequence of acoustic feature vectors, such as log mel filter bank features
or mel frequency cepstral coefficients (MFCC),
recognition proceeds by computing
a weight at every time point for every label,
such as an HMM sub-phonetic state.
The search space is the set of all sub-phonetic state sequences
that corresponds to the set of all word sequences.
Weights for transitioning
from one word to another (the language model) are included
at the states corresponding to boundaries between words.
Recognizing speech becomes the task of finding
the maximum-weight
sequence of states,
considering both the weights from the acoustic features
and the weights from the word transitions.
Since this approach computes a weight for every
acoustic feature vector, or every frame, at every time point,
it is commonly referred to as a frame-based approach.
Many model types that have been proposed as alternatives to HMMs, such as conditional random fields (CRF)~\cite{FHJP2013} and
support vector machine (SVM)-based models~\cite{smith2001speech}, are still frame-based because the search space remains the same.

The inherent limitation of frame-based models is that
the weights can only depend on a fixed length of input at a given time point.
In order to incorporate richer linguistic information,
units other than frames, such as segments~\cite{GZ1986}, have been proposed.
A segment is a variable-length unit,
such as a phoneme~\cite{GZ1986,de2007template} or even a whole word~\cite{BK1985,zweig2009segmental}.
Models operating on segments, known as segmental models, can
take into account the start time, end time, and the associated label
to compute the weights of segments.
The ability to incorporate arbitrary information
within a segment, such as duration~\cite{CS1997}
and acoustic landmarks~\cite{H2005},
makes segmental models appealing for speech recognition.
In fact, segmental models were the state of the art for phoneme recognition
on the TIMIT dataset~\cite{GLFFP1993}
for many years~\cite{glass2003probabilistic}.

However, the flexibility of segmental models comes at a price.
The search space of segmental models includes all possible
ways of segmenting the speech input and all possible ways
of labeling the segments, forming a significantly larger
search space than the one that frame-based models consider.
To bypass the large search space,
early development of segmental models considered 
restricted search spaces produced by pruning based on heuristics~\cite{GZ1986}
or based on a first-pass frame-based recognizer~\cite{glass2003probabilistic,zweig2009segmental}.
Segmental models that operate on
the full search space---first-pass segmental models---have not been
explored until recently~\cite{zweig2012classification}.
Since then, there has been a variety of work exploring
better segment representations for first-pass segmental models,
especially ones that depend on
neural networks~\cite{HF2012,abdel2013deep,tang2015discriminative,lu2016segmental}.

Better, but more computationally expensive, segment representations are
of little practical use unless the efficiency of the models is improved.
Therefore, much of the work on improving
segment representations has been tied to specific approaches
for reducing the search space.
For example, segment representations based on multilayer perceptrons (MLP) are used in~\cite{HF2012}, where the search space is reduced by restricting the form of the weight function;
segment representations based on convolutional neural networks and
MLPs are explored in~\cite{tang2015discriminative}, where
the search space is reduced by pruning;
segment representations based on long short-term memory (LSTM) networks
are used in~\cite{lu2016segmental}, where the search space is reduced by
reducing the time resolution.
In this work, we will consider different segment
representations and study how they behave under different search spaces.

Segmental models have been proposed and rediscovered under
different names based on the definitions of the weight functions
and the training losses.  For example, hidden semi-Markov models
are defined in the generative setting~\cite{ostendorf1996};
semi-Markov CRFs~\cite{sarawagi2004semi} were introduced as segmental models
trained with a log loss; segmental structured SVMs refer to
segmental models trained with a hinge loss~\cite{ZG2013}.
Segmental CRFs~\cite{zweig2009segmental} were proposed as segmental models trained with
marginal log loss.
In this work, we will separate the loss functions
from the definition of segmental models and consider different combinations of
segmental weight functions
and losses.

When segmental models are trained with marginal log loss (or another loss that marginalizes over segmentations),
they can be
trained end to end without the need for ground truth segmentations~\cite{lu2016segmental}.
This property is particularly useful when obtaining ground-truth segmentations, such as
alignments at the phonetic level, is expensive or time-consuming.
Though training systems
end to end reduces the amount of human intervention, the
learned representations may not be interpretable,
making it harder to diagnose errors made by end-to-end systems.
In this work,
we will show results comparing segmental models trained
in multiple stages with intermediate supervision
to ones trained end to end.
In this context, we explore two weight functions,
one based on frame classifiers~\cite{HF2012,tang2015discriminative,abdel2013deep},
and one based on segmental
recurrent neural networks~\cite{kong2016segmental,lu2016segmental}.
We will also compare end-to-end frame-based and segmental models
in terms of their
search spaces and loss functions.
Finally, we will use multitask learning as a tool to constrain
and to analyze the learned representations.

\section{Segmental Models}

We consider the problem of sequence prediction,
such as speech recognition, as a graph search problem.
The graph, usually represented as a finite-state transducer (FST),
is a search space consisting of
all of the ways of segmenting and labeling the input.
A vertex in the graph corresponds to a point in time,
and an edge in the graph corresponds to a segment, that is a time span in the acoustic input and a possible label.
The graph is weighted, and the weight of an edge corresponds
to how well the edge (segment) matches the input.
To compute the weight of an edge, we first
transform the input to an intermediate representation
with a feature encoder.  There are many choices for the type of encoder; here we mainly consider ones based on long short-term memory (LSTM) networks~\cite{hochreiter1997long}.
The intermediate representation is then used
to compute weights, and we refer to the
weighted graph
and the graph search algorithm
as the sequence decoder.
Below we will formally define these components.

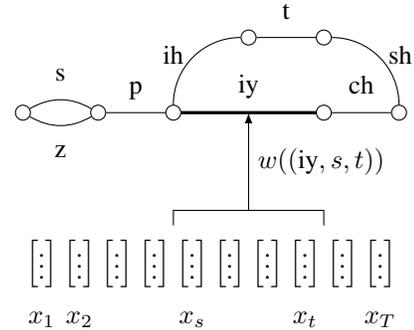
\begin{figure}
\begin{center}
\begin{tikzpicture}[ver/.style={circle,draw,inner sep=2pt}]

\node[ver] (x0) at (0, 0) {};
\node[ver] (x1) at (1, 0) {};
\node[ver] (x2) at (2, 0) {};
\node[ver] (x3) at (4, 0) {};
\node[ver] (x4) at (5, 0) {};

\node[ver] (x5) at (3, 1) {};
\node[ver] (x6) at (4, 1) {};

\draw (x0) edge[out=30, in=150] node [above, font=\strut] {s} (x1);
\draw (x1) edge node [above, font=\strut] {p} (x2);
\draw (x2) edge[very thick] node [above, font=\strut] {iy} (x3);
\draw (x3) edge node [above, font=\strut] {ch} (x4);

\draw (x0) edge[out=-30, in=210] node [below, font=\strut] {z} (x1);
\draw (x2) edge[out=90, in=180] node [left, font=\strut] {ih} (x5);
\draw (x5) edge node [above, font=\strut] {t} (x6);
\draw (x6) edge[out=0, in=90] node [right, font=\strut] {sh} (x4);

\node at ( 0.25, -2) {$\begin{bmatrix}\vdots\end{bmatrix}$};
\node at ( 0.75, -2) {$\begin{bmatrix}\vdots\end{bmatrix}$};
\node at ( 1.25, -2) {$\begin{bmatrix}\vdots\end{bmatrix}$};
\node at ( 1.75, -2) {$\begin{bmatrix}\vdots\end{bmatrix}$};
\node at ( 2.25, -2) {$\begin{bmatrix}\vdots\end{bmatrix}$};
\node at ( 2.75, -2) {$\begin{bmatrix}\vdots\end{bmatrix}$};
\node at ( 3.25, -2) {$\begin{bmatrix}\vdots\end{bmatrix}$};
\node at ( 3.75, -2) {$\begin{bmatrix}\vdots\end{bmatrix}$};
\node at ( 4.25, -2) {$\begin{bmatrix}\vdots\end{bmatrix}$};
\node at ( 4.75, -2) {$\begin{bmatrix}\vdots\end{bmatrix}$};

\node at ( 0.25, -2.75) {$x_1$};
\node at ( 0.75, -2.75) {$x_2$};
\node at ( 1.25, -2.75) {     };
\node at ( 1.75, -2.75) {     };
\node at ( 2.25, -2.75) {$x_s$};
\node at ( 2.75, -2.75) {     };
\node at ( 3.25, -2.75) {     };
\node at ( 3.75, -2.75) {$x_t$};
\node at ( 4.25, -2.75) {     };
\node at ( 4.75, -2.75) {$x_T$};

\draw (2, -1.5) -- (2, -1.3) -- (4, -1.3) -- (4, -1.5);
\draw[->] (3, -1.3) to node [right] {$w((\text{iy}, s, t))$} (3, 0);

\end{tikzpicture}
\caption{An example of a segmental model. A search space is built based on the
    input frames. Each edge (segment) has a start time, an end time, and a label,
    which the weight function can make use of. Once the weights of edges
    are computed, decoding becomes the problem of finding the maximum-weight path.
}
\label{fig:seg}
\end{center}
\end{figure}

Let $\mathcal{X}$ be the input space,
the set of all sequences of acoustic feature vectors,
e.g., log mel filter bank features or
mel frequency cepstral coefficients (MFCCs).
Specifically, for a sequence of $T$ vectors $x = (x_1, \dots, x_T) \in \mathcal{X}$,
each $x_t \in \mathbb{R}^d$, for $t \in \{1, \dots, T\}$,
is a $d$-dimensional acoustic feature vector,
also referred to as a \term{frame}.
Let $\mathcal{Y}$ be the output space,
the set of all label sequences,
where each label in a label sequence comes
from a label set $L$, e.g., a phoneme set in the case
of phoneme recognition.
Given any $T$ frames,
a \term{segmentation} of length $K$ is a sequence of
time points $((0 = s_1, t_1), \dots, (s_K, t_K = T))$,
where $s_k < t_k$ and $t_k = s_{k+1}$ for $k \in \{1, \dots, K\}$.
A \term{segment} (typically denoted $e$ in later sections) is a tuple $(\ell, s, t)$ where
$\ell \in L$ is its label, $s$ is the start time,
and $t$ is the end time.

A \term{segmental model} is a tuple $(\Theta, w)$
where $\Theta$ is a set of parameters,
and $w: \mathcal{X} \times E \to \mathbb{R}$ is a weight function
parameterized by $\Theta$ and $E$ is the set of all segment tuples $(\ell, s, t)$.
The set of parameters $\Theta$ includes all parameters
for both the feature encoder and the sequence decoder.
A sequence of segments forms a \term{path}.
Specifically, a path of length $K$ is a sequence of segments
$(e_1, \dots, e_K)$, where $e_k \in E$ for $k \in \{1, \dots, K\}$.
Let $\mathcal{P}$ be the set of all paths.
For any path $p$, we overload $w$ such that
$w(x, p) = \sum_{e \in p} w(x, e)$.
We will also abbreviate $w(x, e)$ and $w(x, p)$
as $w(e)$ and $w(p)$ respectively when the context
is clear.
An example is shown in Figure~\ref{fig:seg}.

Given an input $x \in \mathcal{X}$,
segmental models aim to solve sequence prediction
by reducing it to finding the maximum-weight path
\begin{equation} \label{eq:inf}
\argmax_{p \in \mathcal{P}} w(x, p).
\end{equation}
The set of paths $\mathcal{P}$, commonly
referred to as the \term{search space},
can be compactly represented as an FST.

Here we briefly review the definition of FSTs.
A multigraph (a graph that can have multiple edges between any pair of vertices)
$G$ is a tuple $(V, E, \tail, \head)$,
where $V$ is a set of vertices,
$E$ is a set of edges,
$\tail: E \to V$ is a function that returns
the vertex where an edge starts,
and $\head: E \to V$ is a function that
returns the vertex where an edge ends.
We deliberately overload $E$,
because every segment has
a corresponding edge in the graph.
An FST is a tuple $(G, \Sigma, \Lambda, I, F, i, o, w)$,
where $G$ is a multigraph,
$\Sigma$ is a set of input symbols,
$\Lambda$ is a set of output symbols,
$I \subseteq V$ is a set of initial vertices,
$F \subseteq V$ is a set of final vertices,
$i: E \to \Sigma$ is a function that defines
the symbol an edge takes as input,
$o: E \to \Lambda$ is a function that
defines the symbol an edge outputs,
and $w: E \to \mathbb{R}$ is
a function that puts weights on edges.
We deliberately overload $w$ as well, because
the weight of a segment will be
the weight of the corresponding edge.

We also associate a time function $\tau: V \to \mathbb{N}$
that maps a vertex to a time point (frame index).
For convenience, we define $\inedges(v) = \{e \in E : \head(e) = v\}$
and $\outedges(v) = \{e \in E: \tail(e) = v\}$.
We will also assume there is one unique start vertex and
unique end vertex, i.e., $|I| = |F| = 1$,
but this can be easily relaxed.
More detailed
discussion of FSTs and their applications for speech
recognition can be found in~\cite{M1997}.

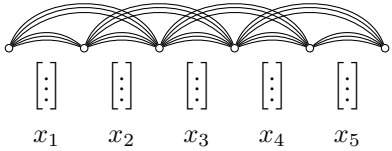
\begin{figure}
\begin{center}
\begin{tikzpicture}[ver/.style={circle,draw,inner sep=1pt}]
\node [ver] (x0) at (0, 0) {};
\node [ver] (x1) at (1, 0) {};
\node [ver] (x2) at (2, 0) {};
\node [ver] (x3) at (3, 0) {};
\node [ver] (x4) at (4, 0) {};
\node [ver] (x5) at (5, 0) {};

\draw (x0) edge [out=20,in=160] (x1);
\draw (x1) edge [out=20,in=160] (x2);
\draw (x2) edge [out=20,in=160] (x3);
\draw (x3) edge [out=20,in=160] (x4);
\draw (x4) edge [out=20,in=160] (x5);

\draw (x0) edge [out=30,in=150] (x1);
\draw (x1) edge [out=30,in=150] (x2);
\draw (x2) edge [out=30,in=150] (x3);
\draw (x3) edge [out=30,in=150] (x4);
\draw (x4) edge [out=30,in=150] (x5);

\draw (x0) edge [out=40,in=140] (x1);
\draw (x1) edge [out=40,in=140] (x2);
\draw (x2) edge [out=40,in=140] (x3);
\draw (x3) edge [out=40,in=140] (x4);
\draw (x4) edge [out=40,in=140] (x5);

\draw (x0) edge [out=50,in=130] (x2);
\draw (x1) edge [out=50,in=130] (x3);
\draw (x2) edge [out=50,in=130] (x4);
\draw (x3) edge [out=50,in=130] (x5);

\draw (x0) edge [out=60,in=120] (x2);
\draw (x1) edge [out=60,in=120] (x3);
\draw (x2) edge [out=60,in=120] (x4);
\draw (x3) edge [out=60,in=120] (x5);

\draw (x0) edge [out=70,in=110] (x2);
\draw (x1) edge [out=70,in=110] (x3);
\draw (x2) edge [out=70,in=110] (x4);
\draw (x3) edge [out=70,in=110] (x5);

\node (y0) at (0.5, -0.5) {$\begin{bmatrix}\vdots\end{bmatrix}$};
\node (y1) at (1.5, -0.5) {$\begin{bmatrix}\vdots\end{bmatrix}$};
\node (y2) at (2.5, -0.5) {$\begin{bmatrix}\vdots\end{bmatrix}$};
\node (y3) at (3.5, -0.5) {$\begin{bmatrix}\vdots\end{bmatrix}$};
\node (y4) at (4.5, -0.5) {$\begin{bmatrix}\vdots\end{bmatrix}$};

\node at (0.5, -1.2) {$x_1$};
\node at (1.5, -1.2) {$x_2$};
\node at (2.5, -1.2) {$x_3$};
\node at (3.5, -1.2) {$x_4$};
\node at (4.5, -1.2) {$x_5$};

\end{tikzpicture}
\caption{An example of a search space for a five-frame utterance
    with a label set $L$ of size three and maximum segment duration $D$ of two frames,
    i.e., $T = 5$, $|L| = 3$, and $D = 2$.
    The three edges between any two nodes are associated with the three labels.
}
\label{fig:search-space}
\end{center}
\end{figure}

To represent the set of paths $\mathcal{P}$
as an FST, we place a vertex at every time point
and connect vertices based on the set of segments.
Specifically, suppose we have $T$ frames.
The set of segments $E$ is an exhaustive enumeration
of tuples $(\ell, s, t)$ for all $\ell \in L$ and
$0 \leq s < t \leq T$.
In practice, a maximum duration $D$ is typically imposed,
i.e., for any segment $(\ell, s, t)$, $t - s \leq D$,
reducing the possible number of segments from $O(T^2|L|)$ to $O(TD|L|)$.
We create a set of vertices $V = \{v_0, v_1, \dots, v_T\}$
such that $\tau(v_t) = t$ for $t \in \{0, 1, \dots, T\}$.
For every segment $(\ell, s, t) \in E$, we create
an edge $e$ such that $i(e) = o(e) = \ell$,
$\tail(e) = v_s$, and $\head(e) = v_t$.
We set $\Sigma = \Lambda = L$, $I = \{v_0\}$, and $F = \{v_T\}$
to complete the construction of the FST
given any $T$ frames.  An example of a search space
is shown in Figure~\ref{fig:search-space}.
One of the many benefits of representing the search space as an FST is that
higher-order segmental models can be constructed by structurally composing
the search space with higher-order language models \cite{tang2015discriminative}.

Given the search space constructed above,
inference, i.e., finding the maximum-weight path \eqref{eq:inf},
can be done efficiently with dynamic programming.
Let $\mathcal{P}(u, v)$ be the set of paths
that starts at vertex $u$ and ends at vertex $v$.
By our previous definition, $\mathcal{P} = \mathcal{P}(v_0, v_T)$.
Define
\begin{equation}
d(v) = \max_{p \in \mathcal{P}(v_0, v)} \sum_{e \in p} w(e).
\end{equation}
In words, $d(v)$ is the maximum path weight
for all paths between $v_0$ and $v$,
and the goal is to find $d(v_T)$.
By definition, we have
\begin{align}
d(v) & = \max_{e \in \inedges(v)} \max_{p' \in \mathcal{P}(v_0, \tail(e))} \left[ w(e)
        + \sum_{e' \in p'} w(e') \right] \\
    & = \max_{e \in \inedges(v)} \Big[ w(e) + d(\tail(e)) \Big].
\end{align}
Algorithm~\ref{alg:inf} shows how to compute all
of the entries in $d$ based on the above recursion
and how to backtrack to find the path;
it is
the same as the shortest path algorithm
for directed acyclic graphs~\cite{M2002}.
Since $v_0, \dots, v_T$ follows a topological order,
Algorithm~\ref{alg:inf} is guaranteed to return the maximum-weight path.

\begin{algorithm}
\caption{Finding the maximum-weight path} \label{alg:inf}
\begin{algorithmic}
\State $d(v_0) = 0$
\For{$v = v_0, v_1, \dots, v_T$}
    \State $d(v) = \max_{e \in \inedges(v)} \Big[ d(\tail(e)) + w(e) \Big]$
    \State $\delta(v) = \argmax_{e \in \inedges(v)} \Big[ d(\tail(e)) + w(e) \Big]$
\EndFor
\State $u = v_T$, $p = \emptyset$
\While{$u \neq v_0$}
    \State $p = \delta(u) \cup p$
    \State $u = \tail(\delta(u))$
\EndWhile
\State \Return $p$
\end{algorithmic}
\end{algorithm}

Since for any vertex $v$, $|\inedges(v)| \leq D|L|$,
the runtime of Algorithm~\ref{alg:inf} is $O(TD|L|) = O(|E|)$.
In fact, Algorithm~\ref{alg:inf} evaluates $w(e)$
for every edge $e \in E$ exactly once.

A segmental model can be trained by finding
a set of parameters $\Theta$ that minimizes a loss
function $\mathcal{L}$.
The model definition
is not tied to any loss function,
allowing us to study the behavior of segmental models
under different loss functions.

\section{Weight Functions}

Here we detail two types of neural weight
functions based on prior work by ourselves and others.
The term feature function is often used in the literature
to denote the function $\phi: \mathcal{X} \times E \to \mathbb{R}^m$
for some $m$, where the weight function $w(x, e)$ is of the form
$\theta^\top \phi(x, e)$ for some parameter vector $\theta \in \mathbb{R}^m$.
When considering neural networks, the weight function
need not be a dot product, but can be any differentiable real-valued
function.

The first type of weight function is similar to those
of~\cite{HF2012,tang2015discriminative,abdel2013deep}, consisting
of outputs of frame-level neural network classifiers ``summarized''
in various ways over the span of a segment; the specific formulation
we use is that of~\cite{tang2016end}.  The second
type of weight function is a segmental recurrent neural network, as in
\cite{lu2016segmental}.  We study segmental models in the context of
these particular weight functions due to their prior success.  In both
cases the acoustic encoder is based on long short-term memory networks
(LSTMs)~\cite{hochreiter1997long}.

Recall that the weight function $w$ takes
a sequence of acoustic features  $x = (x_1, \dots, x_T)$
and a segment $(\ell, s, t)$ as input.
The weight function first passes the acoustic
features through multiple layers of LSTMs.
Let $h_1, \dots, h_{\tilde{T}}$ be the sequence of output
vectors\footnote{Specifically, $h_t = W_f h^f_t + W_b h^b_t$,
where $h^f_t$ and $h^b_t$ are the output vectors
of the forward and backward LSTMs for some weight matrices $W_f$ and $W_b$.
The output vector for the forward LSTM is defined as $h^f_t = \tanh(c^f_t) \odot o^f_t$,
where $c^f_t$ is the cell, and $o^f_t$ is the output gate at time $t$.
The output vector $h^b_t$ is defined similarly for the backward LSTM.
\htcomment{added a more detailed description}
\klcomment{remind me, did we ever compare this averaging of forward and backward states to concatenation (with appropriately resized projection matrix)?}
\htcomment{aren't they equivalent? we are essentially doing
$h_t = \begin{pmatrix} W_f & W_b \end{pmatrix} \begin{pmatrix} h_t^f \\ h_t^b \end{pmatrix}$.}
}
of the final LSTM.
The output of each layer can be subsampled before feeding to the
next layer to reduce the time resolution.
For example, if we subsample at layers two and three for a 3-layer LSTM,
then $\tilde{T} = T/4$.
Otherwise, $\tilde{T} = T$. 
We will use $\Theta_\text{enc}$ to denote the parameters of
the LSTMs, and let $\Theta_\text{dec}$ be
the remaining parameters in the weight function.
Note that $\Theta = \Theta_\text{enc} \cup \Theta_\text{dec}$.

\subsection{FC weight function}

The first type of weight function,
termed the frame classifier (FC) weight,
is similar to weight functions used in a variety of prior work~\cite{HF2012,tang2015discriminative,abdel2013deep}.
A frame classifier takes
in the LSTM output $h_1, \dots, h_{\tilde{T}}$ and produces
a sequence of log probability vectors over the labels
\begin{equation}
z_i = \text{logsoftmax}(W h_i + b)
\end{equation}
where $z_i \in \mathbb{R}^{|L|}$ and $W$ and $b$ are the parameters,
for $i \in \{1, \dots \tilde{T}\}$.  Based on these posterior vectors, we define several functions that summarize the posteriors over a segment:

\paragraph{frame average} The average of transformed log probabilities
\begin{equation}
w_{\text{avg}}((\ell, s, t)) = \frac{1}{t - s} \sum_{i=s}^{t-1} u_{i, \ell},
\end{equation}
where $u_i = W_\text{avg} z_i$ for $i \in \{1, \dots, \tilde{T}\}$.
\paragraph{frame samples} A sample of transformed log probabilities
\begin{equation}
w_{\text{spl-}j}((\ell, s, t)) = u_{j, \ell}
\end{equation}
at time $j \in \{(t - s)/6, (t - s)/2, 5(t - s)/6\}$,
where $u_i = W_\text{spl} z_i$ for $i \in \{1, \dots, \tilde{T}\}$.
\paragraph{boundary} The samples of transformed log probabilities
around the left boundary (start) and right boundary (end) of the segment:
\begin{align}
  w_{\text{left-$k$}}((\ell, s, t)) &= u_{k, i-k, \ell} \\
  w_{\text{right-$k$}}((\ell, s, t)) &= u'_{k, i+k, \ell}
\end{align}
where $u_{k, i} = W_{\text{left-$k$}} z_i$ and
$u'_{k, i} = W_{\text{right-$k$}} z_i$ for $k = 1, 2, 3$
and $i \in \{1, \dots, \tilde{T}\}$.
\paragraph{duration} The label-dependent duration weight
\begin{equation}
w_{\text{dur}}((\ell, s, t)) = d_{\ell, t - s}.
\end{equation}
\paragraph{bias} A label-dependent bias
\begin{equation}
w_{\text{bias}}((\ell, s, t)) = b'_{\ell}.
\end{equation}
The final FC weight function is the sum
of all of the above weight functions.
When the FC weight function is used,
$\Theta_\text{dec}$ is $\{W, b, W_\text{avg}, W_\text{spl},
W_\text{left}, W_\text{right}, d, b'\}$.

\subsection{SRNN weight function}

The second type of weight function is based on segmental recurrent neural networks (SRNNs)~\cite{kong2016segmental,lu2016segmental}.
Suppose the LSTM output vectors are $h_1, \dots, h_{\tilde{T}}$.
To compute $w((\ell, s, t))$, two hidden layers
\begin{align*}
z^{(1)}_{\ell, s, t} & = \text{ReLU}(W_1 [ h_s; h_t; c_\ell; d_k] + b_1) \\
z^{(2)}_{\ell, s, t} & = \tanh(W_2 z^{(1)}_{\ell, s, t} + b_2)
\end{align*}
are computed directly from the LSTM outputs
before computing the final weight,
where $c_\ell$ is a label embedding vector for the label $\ell$,
$d_k$ is a duration embedding vector for the
duration $k$ in log scale,
and $\text{ReLU}(x) = \max(x, 0)$.
The final weight for the segment is defined as
\begin{align*}
w((\ell, s, t)) & = \theta^\top z^{(2)}_{\ell, s, t}.
\end{align*}
Note that instead of encoding the LSTM output vectors
$h_1, \dots, h_{\tilde{T}}$ with an additional LSTM
per segment as in \cite{kong2016segmental}, for efficiency
we use the left and right output vectors $h_s$ and $h_t$
and use a simple feed-forward network
to compute the weight $w((\ell, s, t))$.
When the SRNN weight function is used,
$\Theta_\text{dec}$ is $\{W_1, b_1, W_2, b_2, \theta\}$.
Although the SRNN weight function is conceptually simple,
it is more expensive to compute
than the FC weight function.

\section{Losses}

Recall that a path $p = ((\ell_1, s_1, t_1), \dots, (\ell_K, s_K, t_K))$
consists of a label sequence $y = (\ell_1, \dots, \ell_K)$
and a segmentation $z = ((s_1, t_1), \dots, (s_K, t_K))$.
We will use $(y, z)$ and $p$ interchangeably.
We will denote the space of all segmentations $\mathcal{Z}$.

Training
aims to find a set of parameters $\Theta$
that minimizes the expected task loss, in our case, the expected edit distance
\begin{equation}
\mathbb{E}_{(x, y) \sim \mathcal{D}}[\edit(y, h_\Theta(x))]
\end{equation}
where $h$ is the inference algorithm Algorithm~\ref{alg:inf}
parameterized with $\Theta$, $\edit$ computes the edit distance
of two sequences,
and the expectation is taken over samples $(x, y) \in \mathcal{X} \times \mathcal{Y}$
drawn from a distribution $\mathcal{D}$.
The edit distance is discrete and therefore difficult to optimize;
instead we minimize the expected loss
\begin{equation} \label{eq:exp-loss}
\mathbb{E}_{(x, y) \sim \mathcal{D}}[\mathcal{L}(\Theta; x, y)],
\end{equation}
where $\mathcal{L}$ is a surrogate loss function.
Some surrogate losses refer to a particular choice of segmentation $z$;
in that case we wish to minimize
\begin{equation} \label{eq:exp-seg-loss}
\mathbb{E}_{(x, y, z) \sim \mathcal{D}'}[\mathcal{L}(\Theta; x, y, z)],
\end{equation}
where $\mathcal{D}'$ is a distribution over
$\mathcal{X} \times \mathcal{Y} \times \mathcal{Z}$.
We will use $\mathcal{D}(y | x)$ and $\mathcal{D}'(y, z | x)$
to denote the conditional distribution of
$y$ and $y, z$, respectively, given the input.
Since the distribution $\mathcal{D}$ is unknown,
we use a training set
$S = \{(x_1, y_1), \dots, (x_n, y_n)\}$\footnote{A training set $S = \{(x_1, y_1, z_1), \dots, (x_n, y_n, z_n)\}$ of size $n$ is needed if we optimize \eqref{eq:exp-seg-loss}
with the approximation being $\frac{1}{n} \sum_{i=1}^n \mathcal{L}(\Theta; x_i, y_i, z_i)$.}
of size $n$ to approximate the expectation and instead minimize
\begin{equation}
\frac{1}{n} \sum_{i=1}^n \mathcal{L}(\Theta; x_i, y_i).
\end{equation}

The connection between the surrogate loss $\mathcal{L}$
and the edit distance depends on the choice of loss.
Below we list the loss functions we consider, along with reasons for using them
and their \mbox{(sub)gradients} with respect to
the weight $w(e)$ for some edge $e$.  The \mbox{(sub)gradients} are used in the first-order methods, such as stochastic gradient descent, that we use for optimization.
We assume that the weight function $w$ is differentiable and
the \mbox{(sub)gradients} with respect to the parameters can be obtained with backpropagation.
Other losses for training segmental models, such as ramp loss and empirical Bayes risk,
are not included here but are treated in~\cite{TGL2014}.
There are interesting connections between loss functions and discriminative training criteria in speech recognition (see, for example, \cite{HDSN2008, MWN2010}).

\subsection{Hinge loss}

Given an utterance $x$ and a ground-truth path $p = (y, z)$,
the hinge loss is defined as
\begin{equation}
\mathcal{L}(\Theta; x, p)
    = \max_{p' \in \mathcal{P}} \left[ \cost(p', p) - w(p) + w(p') \right]
\end{equation}
where $\cost$ is a user-defined, non-negative
cost function.
The connection between the hinge loss and the task loss is through
the cost function.
Suppose $\hat{p} = \argmax_{p \in \mathcal{P}} w(p)$ is
the best-scoring path found by Algorithm~\ref{alg:inf}.
The cost of the inferred path $\hat{p}$
can be upper-bounded by the hinge loss:
\begin{align}
\cost(\hat{p}, p) \leq \cost(\hat{p}, p) - w(p) + w(\hat{p})
    \leq \mathcal{L}(\Theta; x, p).
\end{align}
When the cost function is the edit distance, minimizing
the hinge loss minimizes an upper bound on the edit distance.

The hinge loss
is difficult to
optimize when the cost function is the edit distance.
In practice, the cost function is assumed to be decomposable to allow efficient dynamic programming: 
\begin{equation}
\cost(p', p) = \sum_{e' \in p'} \cost(e', p).
\end{equation}
When the cost is decomposable, the hinge loss can be written as
\begin{align*}
\mathcal{L}(\Theta; x, p)
    & = \max_{p' \in \mathcal{P}} \left[ \sum_{e' \in p'} \cost(e', p)
        - \sum_{e \in p} w(e) + \sum_{e' \in p'} w(e') \right] \\
    & = \max_{p' \in \mathcal{P}} \sum_{e' \in p'} \left[ \cost(e', p)
        + w(e') \right] - \sum_{e \in p} w(e),
\end{align*}
and the $\max$ operator in the first term
can be solved with Algorithm~\ref{alg:inf}
by adding the costs to the weights for all segments.

A subgradient of the hinge loss with respect to $w(e)$ is
\begin{equation}
\frac{\partial\mathcal{L}(\Theta; x, p)}{\partial w(e)}
    = -\ind_{e \in p} + \ind_{e \in \tilde{p}}
\end{equation}
where
\begin{equation}
\tilde{p} = \argmax_{p' \in \mathcal{P}} [\cost(p', p) + w(p')],
\end{equation}
which is the path that maximizes the first term in the hinge loss,
and can be obtained with Algorithm~\ref{alg:inf}
with cost added.

Linear models trained with hinge loss are referred to as support vector machines (SVM),
or as structured SVMs when
applied to structured prediction
problems, e.g., sequence prediction in our case.
Segmental models trained with the hinge loss have been studied by~\cite{ZG2013,TGL2014,tang2015discriminative}.

\subsection{Log loss}

Segmental models can be treated as probabilistic
models by defining probability distributions on
the set of all paths. Specifically, the probability
of a path $p = (y, z)$ is defined as
\begin{equation}
P(y, z | x) = P(p | x) = \frac{1}{Z(x)} \exp(w(x, p))
\end{equation}
where
\begin{equation}
Z(x) = \sum_{p' \in \mathcal{P}} \exp(w(x, p'))
\end{equation}
is the partition function.
Given an input $x$ and a ground-truth path $p$, the log loss is defined as
\begin{equation}
\mathcal{L}(\Theta; x, p) = -\log P(p | x).
\end{equation}
Minimizing the log loss is equivalent to maximizing the conditional likelihood.
In addition, the conditional likelihood can be written as
\begin{align*}
P(y, z | x) & = \mathbb{E}_{(y', z') \sim P(y', z' | x)}[\ind_{(y', z') = (y, z)}] \\
    & = 1 - \mathbb{E}_{(y', z') \sim P(y', z' | x)}[\ind_{(y', z') \neq (y, z)}].
\end{align*}
Therefore, maximizing the conditional likelihood is equivalent to minimizing
the expected zero-one loss
\begin{equation}
\mathbb{E}_{(y', z') \sim P(y', z' | x)}[\ind_{(y', z') \neq (y, z)}],
\end{equation}
where $P(y, z | x)$ is used to approximate $\mathcal{D}'(y, z | x)$.
The use of the log loss can be justified by viewing the expectation above
as an approximation of \eqref{eq:exp-seg-loss}.
Segmental models trained with log loss
have been referred to as semi-Markov CRFs~\cite{sarawagi2004semi}.

Since the weight for the ground-truth path $p$ can be efficiently
computed, we are left with the problem of computing the partition function $Z(x)$.
The partition function can also be computed efficiently with the following
dynamic programming algorithm.
Recall that $\mathcal{P}(u, v)$ is the set of paths
that start at vertex $u$ and end at vertex $v$.
For any vertex $v$, define the forward marginal as
\begin{equation}
\alpha(v) = \log \sum_{p' \in \mathcal{P}(v_0, v)} \exp(w(p')).
\end{equation}
By expanding the edges ending at $v$, we have
\begin{align*}
\alpha(v) &= \log \sum_{p' \in \mathcal{P}(v_0, v)} \exp\left(\sum_{e \in p'} w(e)\right) \\
    &= \log \sum_{e \in \inedges(v)} \sum_{p' \in \mathcal{P}(v_0, \tail(e))}
        \exp\left(w(e) + \sum_{e' \in p'} w(e')\right) \\
    &= \log \sum_{e \in \inedges(v)} \exp\left(w(e) + \alpha(\tail(e))\right)
\end{align*}
Similarly, the backward marginal at $v$ is defined as
\begin{equation}
\beta(v) = \log \sum_{p' \in \mathcal{P}(v, v_T)} \exp(w(p')),
\end{equation}
and has a similar recursive structure.
The complete algorithm is shown in Algorithm~\ref{alg:fb}.
Once all entries in $\alpha$ and $\beta$ are computed,
the log partition function is
\begin{equation}
\log Z(x) = \alpha(v_T) = \beta(v_0).
\end{equation}
We store all of the entries in log space for
numerical stability.

The gradient of the log loss with respect to $w(e)$ is
\begin{align*}
\frac{\partial\mathcal{L}(\Theta; x, p)}{\partial w(e)}
    & = -\ind_{e \in p} + \frac{1}{Z(x)}\sum_{p' \ni e} \exp(w(p')) \\
    & = -\ind_{e \in p} + \exp \Big[ \alpha(\tail(e)) + w(e) \\
    & \qquad {} + \beta(\head(e)) - \log Z(x) \Big],
\end{align*}
which can also be efficiently computed once
the marginals are computed.

\begin{algorithm}
\caption{Computing forward and backward marginals} \label{alg:fb}
\begin{algorithmic}
\State $\alpha(v_0) = 0$
\State $\beta(v_T) = 0$
\State $\text{logadd}(a, b) = \log(\exp(a) + \exp(b))$
\For{$v = v_0, v_1, \dots, v_T$}
    \State $\alpha(v) = \text{logadd}_{e \in \inedges(v)} \Big[ \alpha(\tail(e)) + w(e) \Big]$
\EndFor
\For{$v = v_T, v_{T-1}, \dots, v_0$}
    \State $\beta(v) = \text{logadd}_{e \in \outedges(v)} \Big[ \beta(\head(e)) + w(e) \Big]$
\EndFor
\end{algorithmic}
\end{algorithm}

\subsection{Marginal log loss}

Given an input $x$ and a label sequence $y$,
the marginal log loss is defined as
\begin{equation}
\mathcal{L}(\Theta; x, y) = -\log P(y | x) = -\log \sum_{z \in \mathcal{Z}} P(y, z | x)
\end{equation}
where the segmentation is marginalized compared to log loss.
Following the same argument as for log loss,
the marginal distribution
can be written as
\begin{equation}
P(y | x) = 1 - \mathbb{E}_{y' \sim P(y' | x)}[\ind_{y \neq y'}],
\end{equation}
and maximizing the marginal distribution is equivalent to
minimizing the expected zero-one loss
\begin{equation}
\mathbb{E}_{y' \sim P(y' | x)}[\ind_{y \neq y'}],
\end{equation}
where $P(y | x)$ is used to approximate $\mathcal{D}(y | x)$.
Note that the zero-one loss $\ind_{y \neq y'}$ only depends on the label sequence.
While the log loss has a connection to \eqref{eq:exp-seg-loss},
the marginal log loss
directly approximates \eqref{eq:exp-loss}
with the above expected zero-one loss.

Note that both the hinge and log loss depend on the ground-truth
segmentation.
The marginal log loss does not require the ground-truth
segmentation, making it attractive for
tasks such as speech recognition, because collecting
ground-truth segmentations for phonemes or words is time-consuming
and/or expensive.  In addition, the boundaries of phonemes and words
tend to be ambiguous, so it can be preferable to leave the decision
to the model.
Segmental models trained with the marginal log loss have
been referred to as segmental CRFs~\cite{zweig2009segmental}.

To compute the marginal log loss,
we can rewrite it as
\begin{align}
\mathcal{L}(\Theta; x, y) & = -\log \sum_{z \in \mathcal{Z}} P(y, z | x) \\
    & = -\log \sum_{z \in \mathcal{Z}} \exp(w(x, (y, z))) + \log Z(x) \\
    & = -\underbrace{\log \sum_{p': \Gamma(p') = y} \exp(w(x, p'))}_{\log Z(x, y)} + \log Z(x)
\end{align}
where $\Gamma$ extracts the label sequence from a path, i.e.,
for $p' = (y', z')$, $\Gamma(p') = y'$.
Since the partition function can be efficiently computed
from Algorithm~\ref{alg:fb}, we only need to compute
$\log Z(x, y)$. Since the term $\log Z(x, y)$ is identical to $\log Z(x)$
except that it involves a constrained 
search space considering all paths with the same label sequence $y$,
the strategy is to construct the constrained search space
with an FST and run Algorithm~\ref{alg:fb} on the FST.
Let $F$ be a chain FST that represents $y$,
with edges $\{e_1, \dots, e_{|y|}\}$,
where $i(e_k) = o(e_k) = y_k$ for all $k \in \{1, \dots, |y|\}$.
Let $G$ be the search space consisting of all paths
in $\mathcal{P}$. The term $\log Z(x, y)$
can be efficiently computed by running Algorithm~\ref{alg:fb} on
the intersection of $G$ and $F$, i.e., $G \cap F$.
Let the forward and backward marginals computed on $G \cap F$
be $\alpha'$ and $\beta'$.
We have $\log Z(x, y) = \alpha'(v_T) = \beta'(v_0)$.

The gradient of the marginal log loss is
\begin{align*}
& \frac{\partial\mathcal{L}(\Theta; x, y)}{\partial w(e)} \\
    & = -\frac{1}{Z(x, y)} \sum_{\substack{p' \ni e\\ \Gamma(p') = y}} \exp(w(p'))
        + \frac{1}{Z(x)}\sum_{p' \ni e} \exp(w(p')) \\
    & = - \exp \Big[ \alpha'(\tail(e)) + w(e) + \beta'(\head(e)) - \log Z(x, y) \Big] \\
    & \quad {} + \exp \Big[ \alpha(\tail(e)) + w(e) + \beta(\head(e)) - \log Z(x) \Big].
\end{align*}
and can be efficiently computed once the marginals are computed.

\section{Multi-stage Training and Multitask Training}

Following the conventional ASR pipeline,
we can first build a frame classifier
and use it to build the rest of the segmental models.
Such an approach, using the weight functions defined above,
has been successful
for training segmental models, either
for multi-stage training or as an initialization
for end-to-end training~\cite{tang2016end,lu2017multi}.
We will review these training approaches in detail
and present a unified view for both.

Recall that our parameters can be divided into two parts:
$\Theta_\text{enc}$ for the acoustic feature encoder 
and $\Theta_\text{dec}$ for the sequence decoder.
The acoustic feature encoder
can be trained jointly with the sequence decoder,
or
separately with
other loss functions, such as the frame-wise cross entropy
or the connectionist temporal classification (CTC) loss~\cite{graves2006connectionist}.
We refer to the case where the encoder and decoder are trained jointly
as \term{end-to-end training}, and the case where 
the training is separated into multiple stages (end-to-end or not)
as \term{multi-stage training}.

Consider the end-to-end training approach.  We can write
the objective
\begin{equation} \label{eq:e2e}
\min_{\thetaenc, \thetadec} \mathcal{L}(\thetaenc, \thetadec),
\end{equation}
in terms of both $\thetaenc$ and $\thetadec$
where $\mathcal{L}$ is a loss function that takes
both the encoder and the decoder into account, such as the hinge loss,
log loss, or marginal log loss.
For multi-stage training, we use a loss function
to train the encoder in the first stage by solving
\begin{equation}
\hat{\Theta}_\text{enc} = \argmin_{\thetaenc} \mathcal{L}_\text{enc}(\thetaenc),
\end{equation}
where $\mathcal{L}_\text{enc}$ can be the frame-wise cross entropy
or the CTC loss. In the second stage, we use the obtained $\hat{\Theta}_\text{enc}$
to solve
\begin{equation}
\hat{\Theta}_\text{dec} = \argmin_{\thetadec} \mathcal{L}(\hat{\Theta}_\text{enc}, \thetadec)
\end{equation}
while holding the first argument in the loss fixed.
In the third stage, we can then use $\hat{\Theta}_\text{enc}$ and $\hat{\Theta}_\text{dec}$ as
initialization and solve~\eqref{eq:e2e}.

In addition, we can also consider a convex combination of multiple loss functions
\begin{equation} \label{eq:multitask}
\min_{\thetaenc, \thetadec} \lambda \mathcal{L}(\thetaenc, \thetadec)
    + (1 - \lambda) \mathcal{L}_\text{enc}(\thetaenc)
\end{equation}
where $\lambda$ is the interpolation factor.
End-to-end training can be seen as optimizing \eqref{eq:multitask}
with $\lambda = 1$, while multi-stage training can be seen
as optimizing the second term in \eqref{eq:multitask}
followed by optimizing the first term. 

While there are many benefits for training systems end to end,
such as the potential to find a better optimum and without
requiring supervision at the intermediate level,
end-to-end training might be challenging due to optimization difficulties
and might require more samples.
On the other hand, while multi-stage training requires supervision at
the intermediate level, it might make the optimization easier (sometimes making it convex),
might require fewer samples, and might produce models that are more interpretable.

\section{Experiments}

We apply segmental models to phonetic recognition on TIMIT,
a
dataset \klcomment{edited}
consisting of a training set of 3696 utterances
and a test set, of which a subset of 192 utterances
is called the core test set. Following standard protocol~\cite{P2011},
we use 400 utterances from the complete test set (disjoint from the core test set)
as the validation set, and report the final results on the core test set.
In addition, we reserve 376 utterances from the training set for tuning hyperparameters, such as optimizers, step sizes, and dropout rates,
and use the remaining 3320 utterances for training.
The development set is used solely for early stopping.
As is often done for TIMIT experiments, \klcomment{edited} we collapse the 61 phones in the phone set to 48 for training,
and further collapse them to 39 for evaluation~\cite{L1988}.
TIMIT is phonetically transcribed, so we have the option
of training the feature encoder with frame-wise cross entropy
based on the ground-truth frame labels.
The acoustic input to the feature encoder consists of 40-dimensional log
filter bank features (without energy) and their first and second derivatives.
The resulting 120-dimensional acoustic features are speaker-normalized
by subtracting the per-speaker mean and dividing by the per-speaker standard deviation of
every dimension.

The feature encoder is a 3-layer bidirectional LSTM with 250 hidden units in each direction.
Previous work has shown that subsampling either
the frames or the LSTM outputs can reduce the decoding
time while maintaining accuracy~\cite{VDH2013,MLWZG2015}.
We consider subsampling the output
of the LSTMs by a factor of two after
the second and third layers.
The subsampled
encoder is referred to as a pyramid encoder~\cite{lu2016segmental}.
Dropout~\cite{ZSV2014} is added to
the input and output of the LSTMs at a rate of 0.2.

For the segmental models, we enforce a maximum segment duration of 30 frames
when a regular feature encoder is used, and
a maximum duration of 8 when a pyramid feature encoder is used.
The maximum duration is applied to all labels, including silences.
For the SRNN weight function, following~\cite{lu2016segmental},
the duration embedding is of size 5,
the label embedding is of size 32, and the two subsequence hidden layers
are both of size 64.
All parameters of the weight functions
are initialized based on~\cite{GB2010}.

All loss functions are optimized with stochastic gradient descent (SGD)
with a minibatch size of 1 utterance.
The gradient norm is clipped to 5.
The default optimizer is vanilla SGD unless otherwise stated.
We run the optimizer for 20 epochs with step size 0.1; starting from the best model among the first 20 epochs,
we run for another 20 epochs
with step size decayed by 0.75 after each epoch (i.e., exponential decay).
We choose the epoch that has the best performance
on the development set (early stopping).

\subsection{Multi-stage training}

We first compare different segmental models trained in multiple stages.
The first stage trains the feature encoder $\thetaenc$ either with
the frame-wise cross entropy or with the CTC loss, and the second
stage trains $\thetadec$ with hinge loss, log loss, or marginal log loss.
Finally, after the second stage, we fine-tune both $\thetaenc$ and $\thetadec$
with each of the three losses.

To construct a frame classifier,
the 250-dimensional output vectors
of the encoder are projected down to 48 dimensions
followed by a softmax layer.
Depending on whether we use a pyramid encoder,
we subsample the frame labels accordingly during training.
The resulting frame classifier achieves
frame error rates of 18.3\% for the regular encoder
and 29.1\% for the pyramid encoder (where outputs are upsampled to evaluate performance)
on the development set.

For the CTC loss,
we project the 250-dimensional output vectors of the feature
encoder down to 49 dimensions (48 phones + 1 blank) and
pass them through a softmax layer.
The encoder is fixed to a pyramid, and frame labels are not required during training.
The encoder trained with the CTC loss achieves a phoneme error rate (PER) of
17.2\% on the development set with best-path decoding
(followed by removing duplicates and blanks).

In the first set of experiments, we only use the frame classifiers
as encoders (pyramid or not), and compare the two weight functions.
Since we have the pretrained feature encoders,
we freeze the encoder parameters $\thetaenc$
and train the decoder parameters $\thetadec$.
The default SGD optimizer (20 epochs without decay plus 20 epochs with exponential decay)
is used for the SRNN weight function
because it works well with the two-layer networks in the weight function.
For the FC weight function, note that hinge loss and log loss are
convex in $\thetadec$.
In particular, when the encoder $\thetaenc$ is frozen,
optimizing hinge loss and log loss for the FC weight function
are both convex problems.
RMSprop~\cite{TH2012} is favored over vanilla SGD for the FC weight function with step size $10^{-4}$
and decay 0.9 for 20 epochs.
After the two-stage training, we can further optimize
both the encoder and the decoder.
Here vanilla SGD is used with the step size starting from 0.1
and decayed by 0.75 after each epoch, because
the training loss is already low after two-stage training.

The multi-stage training results are shown in Table~\ref{tbl:weight-comp}.
The results are consistent with those reported in~\cite{tang2015discriminative}.
For the FC weight function with a regular encoder,
the three losses perform equally well,
with marginal log loss having a slight edge over the other two.
Using the pyramid encoder hurts the performance of hinge loss
and log loss, but has less impact on marginal log loss.
Hinge loss and log loss might be more sensitive
to the reduced time resolution because they
are tied to a specific segmentation, while marginal log loss
is more forgiving due to the marginalization.
Fine-tuning improves over two-stage training across all cases.
The conclusion stays the same for the SRNN weight function,
except that training the SRNN weight function without the pyramid is very time-consuming,
and we only manage to complete a few epochs in the two-stage setting.
Although the best results after fine tuning are roughly the same for both weight functions,
to shorten the experimental cycle,
we favor the better performer, the SRNN weight function, with a pyramid encoder in
the two-stage setting. 

\begin{table}
\begin{center}
\caption{Phone error rates (PER, \%) on the development set for segmental models with different weight functions
    and different feature encoders.  The encoders (pyramid or not) are
    trained with the frame-wise cross entropy.
    Two-stage training is denoted \textnormal{2s},
    and two-stage training followed by fine-tuning is
    denoted \textnormal{2s+ft}.
    (* too slow to complete)
}
\label{tbl:weight-comp}
\begin{tabular}{lll|lll}
      & pyramid     &        & hinge & log loss & marginal log loss  \\
\hline
FC
      &             & 2s     & 19.9  & 20.7     & 19.9               \\
      &             & 2s+ft  & 19.3  & 18.2     & 17.9               \\
      & \checkmark  & 2s     & 31.3  & 32.0     & 24.4               \\
      & \checkmark  & 2s+ft  & 23.3  & 22.3     & 17.9               \\
\hline
SRNN
      &             & 2s     & 22.2* & 22.2*    & 20.5*              \\
      &             & 2s+ft  & -     & -        & -                  \\
      & \checkmark  & 2s     & 27.4  & 24.4     & 21.3               \\
      & \checkmark  & 2s+ft  & 24.4  & 22.7     & 18.1               
\end{tabular}
\end{center}
\vspace{-0.2cm}
\end{table}

After fixing the weight function to the SRNN,
we compare encoders pretrained with the frame-wise cross entropy
and with the CTC loss.
Results are shown in Table~\ref{tbl:encoder}.
It is clear that for all losses, using the encoder pretrained
with the CTC loss leads to better performance.

\begin{table}
\begin{center}
\caption{PERs (\%) for segmental models trained in multiple stages with
    feature encoders pretrained with the frame-wise cross
    entropy or the CTC loss. Two-stage training is denoted \textnormal{2s},
    and two-stage training followed by fine-tuning is
    denoted \textnormal{2s+ft.}}
\label{tbl:encoder}
\begin{tabular}{l|lll|lll}
         & \multicolumn{3}{l}{frame loss} & \multicolumn{3}{|l}{CTC} \\
\hline
         & 2s     & 2s+ft & 2s+ft  & 2s      & 2s+ft   & 2s+ft  \\
         & dev    & dev   & test   & dev     & dev     & test   \\
\hline                                                             
hinge    & 27.4   & 24.4  &        & 26.7    & 24.1    &        \\
log loss & 25.9   & 22.7  &        & 25.7    & 22.7    &        \\
marginal log loss                                      
         & 21.3   & 18.1  & 20.9   & 18.7    & 17.8    & 20.2   
\end{tabular}
\end{center}
\vspace{-0.4cm}
\end{table}

\subsection{End-to-end training from random initialization}

In this section, we compare losses for end-to-end
training of segmental models with the pyramid encoder and the SRNN weight function.
Unlike in multi-stage training, all of the experiments here
are trained from random initialization.
Results are shown in Table~\ref{tbl:seg-e2e}.
While the log loss and marginal log loss achieve
reasonable performance, the hinge loss completely fails.
We find that hinge loss values on the training set
are higher compared to the multi-stage models,
suggesting that there is an optimization issue.
Since log loss can be minimized reasonably well,
we suspect that hinge loss is difficult to minimize because
of its non-smoothness.
The result with the marginal log loss is consistent with
reported numbers in previous work \cite{lu2016segmental}.
The performance of CTC is on par
with the segmental model trained with marginal log loss.

\begin{table}
\begin{center}
\caption{Segmental models trained end to end with different losses
    compared with the CTC model.}
\label{tbl:seg-e2e}
\begin{tabular}{lllll}
         & dev   & test  \\
\hline
hinge    & 74.7  &       \\
log loss & 22.2  &       \\
marginal log loss
         & 17.5  & 19.5  \\
\hline
CTC      & 17.2  & 19.5
\end{tabular}
\end{center}
\end{table}

\subsection{End-to-end multitask training}

Instead of optimizing different losses in different stages
as in the previous section,
we next optimize multiple losses jointly from random initialization.
Here we only focus on marginal log loss paired with either
the frame-wise cross entropy or the CTC loss, because the marginal log loss
is the best performer in the previous experiments.
We use early stopping based on the PERs of the segmental model
on the development set.
Results are shown in Table~\ref{tbl:multitask}.
We see that end-to-end training with multiple tasks further improves
over end-to-end training with a single task.  The best test-set result (and the best dev-set result) is obtained by multitask training with marginal log loss + CTC loss, and improves over the CTC error rate by 1\% absolute (19.5\% $\longrightarrow$ 18.5\% on the test set).

\begin{table}
\begin{center}
\caption{PERs (\%) for segmental models trained end to end with multiple tasks,
    i.e., the marginal log loss plus either the frame-wise cross entropy
    or the CTC loss.}
\label{tbl:multitask}
\begin{tabular}{l|ll|ll}
             & \multicolumn{2}{l}{frame loss} & \multicolumn{2}{|l}{CTC} \\
$\lambda$    & dev    & test   & dev    & test   \\
\hline
0.16         & 18.6   &        & 17.5   &        \\
0.33         & 18.0   &        & 17.0   &        \\
0.5          & 17.2   &        & 17.0   &        \\
0.67         & 16.9   &        & 16.7   & 18.5   \\
0.84         & 16.8   & 19.3   & 17.0   &        \\
\hline
1.00         & 17.5   & 19.5   & 17.2   & 19.5   
\end{tabular}
\end{center}
\vspace{-0.4cm}
\end{table}

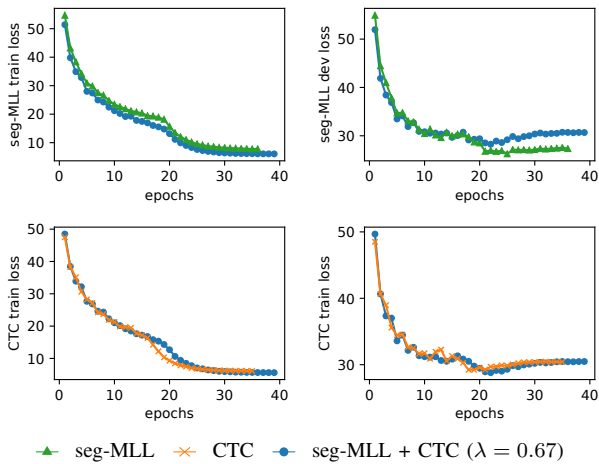
\begin{figure}
\definecolor{c0}{HTML}{1F77B4}
\definecolor{c1}{HTML}{FF7F0E}
\definecolor{c2}{HTML}{2CA02C}
\begin{center}
\includegraphics[width=4cm]{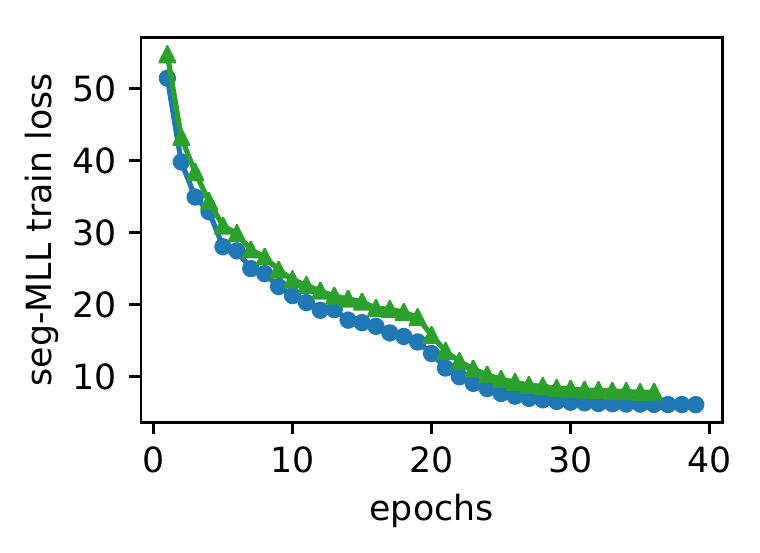}
\includegraphics[width=4cm]{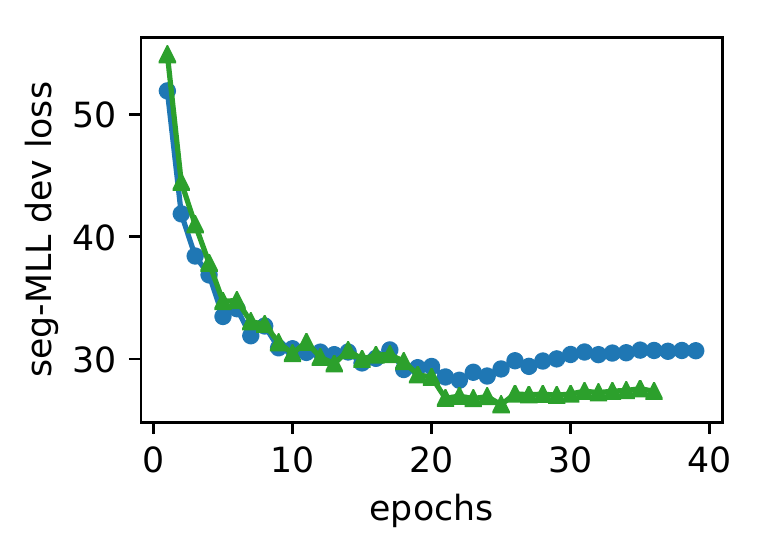}
\includegraphics[width=4cm]{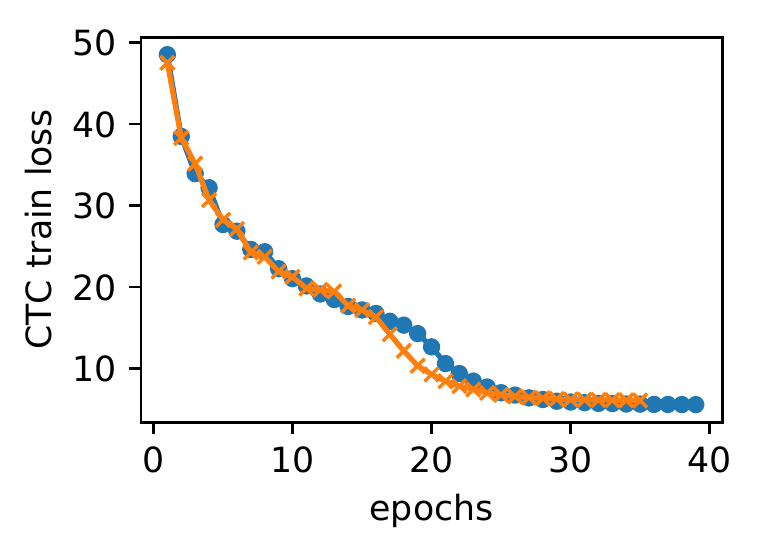}
\includegraphics[width=4cm]{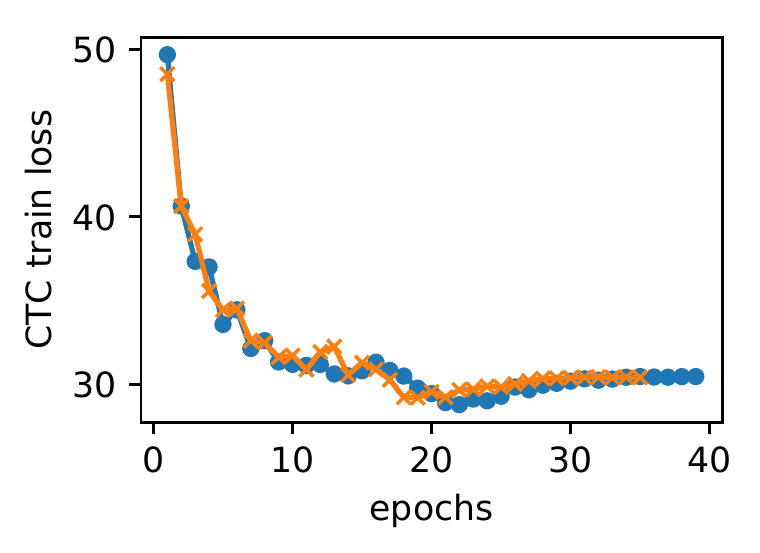}
\begin{tikzpicture}[font=\footnotesize\strut]
\draw[color=c2] (0.1, 0) -- (0.4, 0);
\path[fill=c2] (0.18, -0.04) -- (0.32, -0.04) -- (0.25, 0.08) -- cycle;
\node at (0.5, 0) [right] {seg-MLL};
\end{tikzpicture}
\begin{tikzpicture}[font=\footnotesize\strut]
\draw[color=c1] (0.1, 0) -- (0.4, 0);
\draw[color=c1] (0.18, 0.06) -- (0.32, -0.06);
\draw[color=c1] (0.18, -0.06) -- (0.32, 0.06);
\node at (0.5, 0) [right] {CTC};
\end{tikzpicture}
\begin{tikzpicture}[font=\footnotesize\strut]
\draw[color=c0] (0.1, 0) -- (0.4, 0);
\path[fill=c0] (0.25, 0) circle (0.06);
\node at (0.5, 0) [right] {seg-MLL + CTC ($\lambda=0.67$)};
\end{tikzpicture}
\caption{Loss values for jointly optimizing
    the marginal log loss (seg-MLL) and the CTC loss
    compared with optimizing them individually.}
\label{fig:multitask-loss}
\end{center}
\end{figure}

The success of multitask learning in Table~\ref{tbl:multitask} indicates that
there exists an encoder that can generate representations
suitable for both tasks.  We further investigate the loss values
for the case of jointly optimizing the marginal log loss
and the CTC loss.
The learning curve is shown in Figure~\ref{fig:multitask-loss}.
In the multitask case, both the marginal log loss and
the CTC loss achieve lower values on the training set compared to the single-task case,
suggesting that multitask learning might help optimization.
However, both loss values on the development set end up higher
when multiple losses are used.
The fact that models with higher losses on the development set
end up having lower PERs is unsatisfying
and needs further investigation.

The time to compute gradients for different losses is
shown in Table~\ref{tbl:time}.
All numbers are measured on a single quad-core 3GHz CPU,
averaged over the entire training set.
As a reference, the average real-time factor for
computing the gradients from the outputs of a regular 3-layer LSTM is 0.301,
and reduces to 0.181 for pyramid 3-layer LSTM.
Computing the gradient of the CTC loss is also 0.181.
Though the exact number depends on the implementation,
the general trend is clear.
It is faster to compute the FC weight function than
the SRNN weight function.
Using the pyramid encoder significantly reduces the runtime.
Computing the gradient of hinge loss is fastest,
and computing the gradient of marginal log loss is the slowest.
The real-time factor for decoding is 0.259 including the pyramid LSTM.
Decoding in multiple passes can be an option if speedup is needed~\cite{TWGL2016}.

\begin{table}
\begin{center}
\caption{Average real-time factor per sample
    to compute gradients for different losses.}
\label{tbl:time}
\begin{tabular}{ll|lll}
      & pyramid     & hinge & log loss & marginal log loss  \\
\hline
FC
      &             & 0.257 & 0.811    & 0.954              \\
      & \checkmark  & 0.015 & 0.042    & 0.073              \\
\hline
SRNN
      &             & 3.253 & 5.669    & 6.110              \\
      & \checkmark  & 0.165 & 0.286    & 0.384              
\end{tabular}
\end{center}
\vspace{-0.4cm}
\end{table}

\section{Related Work}

\subsection{First-pass segmental models}

Many models, such as semi-Markov CRFs \cite{sarawagi2004semi},
segmental CRFs \cite{zweig2009segmental}, and inverted HMMs \cite{DHSN2016},
are special cases of segmental models with different weight functions and
trained with different losses.
In Table~\ref{tbl:seg}, we provide a set of highlights of results in the development of segmental models on the TIMIT data set.
Zweig~\cite{zweig2012classification} was the first to explore
discriminative segmental models that search over sequences and segmentations exhaustively,
and did not use neural networks.
He \& Fosler-Lussier~\cite{HF2012} first used (shallow) neural network-based frame classifiers to define weight functions,
and later extended the idea to deep neural networks in~\cite{H2015}.
Abdel-Hamid {\it et al.}~\cite{abdel2013deep} were the first to use
deep convolutional neural networks for the weight functions,
and were the first to train segmental models end to end.
Tang {\it et al.} first compared different
losses and training strategies for segmental models in a rescoring framework~\cite{TGL2014} and then in first-pass segmental models~\cite{tang2016end}.
They also introduced segment-level classifiers and segmental cascades for incorporating them (and other expensive features) into segmental weight functions~\cite{tang2015discriminative}.
Lu {\it et al.}~\cite{lu2016segmental} introduced an LSTM-based weight function for every segment,
and were
also the first to use pyramid LSTMs to speed up inference for segmental models.

\begin{table}
\begin{center}
\caption{TIMIT PERs (\%) for various segmental models
    compared with HMMs and the state of the art.
    The acoustic features can be speaker independent (\textnormal{spk indep}) or
    speaker adapted with mean and variance normalization (\textnormal{mvn})
    or maximum likelihood linear regression (\textnormal{fMLLR})~\cite{P2011}.  Some results were obtained with MFCCs and some with log filter bank features.
}
\label{tbl:seg}
\begin{tabular}{lp{0.65cm}p{0.6cm}p{1.1cm}}
              & spk indep   & +mvn       & +fMLLR \\
\hline
HMM-DNN \cite{P2011}
              & 21.4        &             & 18.3 \\
HMM-CNN \cite{T2015}
              & 16.5 \\
\hline
SUMMIT (1998) \cite{HG1998,glass2003probabilistic}
              & 24.4 \\
segmental CRF (SCRF) (2012) \cite{zweig2012classification}
              & 33.1 \\
SCRF + shallow NN (2012) \cite{HF2012}
              & 26.5 \\
SCRF + DNN (2015) \cite{H2015}
              &             &             & 19.1 \\
deep segmental NN (2013) \cite{abdel2013deep}
              & 21.9 \\
segmental cascades
(2015) \cite{tang2015discriminative}
              & 19.9 \\
segmental RNN (SRNN) (2016) \cite{lu2016segmental}
              &             & 18.9        & 17.3 \\
end-to-end + two-stage training (2016) \cite{tang2016end}
              & 19.7 \\
SRNN + multitask (2017) \cite{lu2017multi}
              &             & 18.7        & 17.5 \\
SRNN + multitask (2017) (this work)
              &             & 18.5
\end{tabular}
\end{center}
\vspace{-0.4cm}
\end{table}

\subsection{End-to-end models}

Most mainstream end-to-end speech recognition models
can be broadly categorized as either frame-based models or
encoder-decoder models.
CTC, HMMs, and some newer approaches like
the auto-segmentation criterion (ASG)~\cite{CPS2016}
fall under the first category, because these models
emit one symbol for every frame. 
Falling under the second category, encoder-decoder
models proposed by~\cite{CBSCB2015,BCSBB2016,CJLV2016}
generate labels one at a time while
conditioning on the input and the labels generated
in the past, without an explicit alignment between labels and frames.
Since frame-based models follow the same
graph search framework as segmental models, we will
focus on discussing the connection between these and segmental models.

Recall that training segmental models with marginal log
loss requires a search space $G$, a constraint FST $F$
to limit the search space to ground-truth labels, and the loss itself.
To compute marginal log loss, we first compute the marginals
on $G$ for computing the partition function $Z(x)$,
and then compute the marginals on the intersection $G \cap F$
for computing $Z(x, y)$.
CTC, HMMs, and ASG can all be seen as special cases
of this framework.

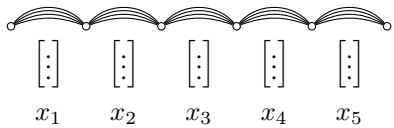
\begin{figure}
\begin{center}
\begin{tikzpicture}[ver/.style={circle,draw,inner sep=1pt}]
\node [ver] (x0) at (0, 0) {};
\node [ver] (x1) at (1, 0) {};
\node [ver] (x2) at (2, 0) {};
\node [ver] (x3) at (3, 0) {};
\node [ver] (x4) at (4, 0) {};
\node [ver] (x5) at (5, 0) {};

\draw (x0) edge [out=20,in=160] (x1);
\draw (x1) edge [out=20,in=160] (x2);
\draw (x2) edge [out=20,in=160] (x3);
\draw (x3) edge [out=20,in=160] (x4);
\draw (x4) edge [out=20,in=160] (x5);

\draw (x0) edge [out=30,in=150] (x1);
\draw (x1) edge [out=30,in=150] (x2);
\draw (x2) edge [out=30,in=150] (x3);
\draw (x3) edge [out=30,in=150] (x4);
\draw (x4) edge [out=30,in=150] (x5);

\draw (x0) edge [out=40,in=140] (x1);
\draw (x1) edge [out=40,in=140] (x2);
\draw (x2) edge [out=40,in=140] (x3);
\draw (x3) edge [out=40,in=140] (x4);
\draw (x4) edge [out=40,in=140] (x5);

\draw (x0) edge [out=50,in=130] (x1);
\draw (x1) edge [out=50,in=130] (x2);
\draw (x2) edge [out=50,in=130] (x3);
\draw (x3) edge [out=50,in=130] (x4);
\draw (x4) edge [out=50,in=130] (x5);

\node (y0) at (0.5, -0.5) {$\begin{bmatrix}\vdots\end{bmatrix}$};
\node (y1) at (1.5, -0.5) {$\begin{bmatrix}\vdots\end{bmatrix}$};
\node (y2) at (2.5, -0.5) {$\begin{bmatrix}\vdots\end{bmatrix}$};
\node (y3) at (3.5, -0.5) {$\begin{bmatrix}\vdots\end{bmatrix}$};
\node (y4) at (4.5, -0.5) {$\begin{bmatrix}\vdots\end{bmatrix}$};

\node at (0.5, -1.2) {$x_1$};
\node at (1.5, -1.2) {$x_2$};
\node at (2.5, -1.2) {$x_3$};
\node at (3.5, -1.2) {$x_4$};
\node at (4.5, -1.2) {$x_5$};

\end{tikzpicture}
\caption{An example of the CTC search space for a five-frame utterance
    with a label set of size three (plus one blank). 
}
\label{fig:ctc}
\end{center}
\vspace{-0.4cm}
\end{figure}

The search space of CTC
has an edge
for every label in the label set (including the blank label)
at every time step.  Specifically, the search space $G$ includes
the edges $\{e_{\ell, t}: \ell \in L, t \in \{1, \dots, T\}\}$
with $v_{t-1} = \tail(e_{\ell, t})$ and $v_t = \head(e_{\ell, t})$.
An example is shown in Figure~\ref{fig:ctc}.
The weight of an edge $e_{\ell, t}$ is the log probability
of label $\ell$ at time $t$.  By construction, the decision
made at every time point is independent of the decision
at other time points.  In addition, since the probabilities
at every time point sum to one, the partition function $Z(x)$
of the search space is always $1$.  The constraint FST $F$
representing the ground-truth labels consists of the sequences of one or more labels
with zero or more blanks in between labels.
For example, for the label sequence ``k ae t,''
the constraint FST is the regular expression
$\varnothing^*\text{k}^+\varnothing^*\text{ae}^+\varnothing^*\text{t}^+\varnothing^*$.
With the above construction, marginal log loss becomes
exactly the objective of CTC.

Comparing CTC to HMMs, the search space is different depending
on the HMM topology.
For example, two-state HMMs are used in~\cite{povey2016purely}.
Since the transition probabilities and posterior probabilities
are all locally normalized, the partition function $Z(x)$
is always $1$.  The constraint FST representing the ground-truth labels
consists simply of sequences of repeating labels.
For example, for the label sequence ``k ae t,''
the constraint FST is the regular expression $\text{k}^+\text{ae}^+\text{t}^+$.
With the above construction, marginal log loss applied to HMMs is equivalent
to lattice-free MMI~\cite{povey2016purely}.

For ASG, the search space is equivalent to that of one-state HMMs.
Instead of assuming conditional independence as in CTC,
ASG includes transition probabilities between states.
The constraint FST is identical to that of HMMs, with repeated
ground-truth labels. However, in ASG
the weights on the edges
are not locally normalized, so
the partition function $Z(x)$ is not always $1$ and has to be
computed.
With the above search space construction, marginal log loss
becomes ASG.

Another approach similar to CTC proposed in~\cite{G2012}
is called RNN transducers.  The search space of
an RNN transducer is the set of alignments from the speech
signal to all possible label sequences,
so the search space grows exponentially in the number of labels.
The weight function of a path in this approach relies
on an RNN, and is not decomposable as a sum of weights of the edges.
RNN transducers are trained with
marginal log loss.
By the independence assumption imposed in~\cite{G2012},
the partition function $Z(x)$ is still $1$,
so we do not need to marginalize over the exponentially
large space.  During decoding, however, we still have to search
over the exponentially large space with, for example, beam search.

In view of this framework, even when using the same loss function,
i.e., marginal log loss,
segmental models and frame-based models differ in their search space,
weight functions, and how the search space is constrained by the ground truth labels during training.

\subsection{Word recognition}

First-pass segmental models have previously been successfully applied to word recognition \cite{glass2003probabilistic,he2015segmental}.
This previous work treats first-pass segmental models as a drop-in replacement
for HMM phoneme recognizers, because both models serve as
functions that map acoustic features to phoneme strings.
The phoneme recognizers are then composed with a lexicon and a language model
to form a word recognizer.

Recent work has explored models that directly predict characters, avoiding the need for a lexicon~\cite{graves2014towards,MXJN2015,miao2015eesen} but still allowing for improved performance when constraining the search space
with a lexicon (through FST composition)~\cite{miao2015eesen}.
Segmental models can also be used to predict characters
simply by changing the label set.

Instead of using intermediate discrete representations, such as
phonemes or characters, recent advances in computing power have made it feasible to directly predict words~\cite{MMONN2012,BH2014,SLS2016,ARSPN2017}.
In this case, rather than using a pronunciation dictionary,
only a list of words is needed for decoding.
Segmental models can also be used to directly predict
words by using the list of words as the label set.
This approach is worth exploring further, although efficiency issues
make it nontrivial to train such models~\cite{ARSPN2017}.

\section{Conclusion}

We have presented the formal framework of segmental models
and several potential losses for training such models.  Segmental models are now able to run efficiently enough for end-to-end training and obtain competitive error rates.
We have explored segmental models with two types of weight functions and various training losses on the task of phonetic recognition.  We have found that the best results obtained with the two types of weight functions (frame classifier-based and segmental recurrent neural networks) are quite similar, and are typically best with marginal log loss.

We also consider the relationship between segmental models and
frame-based models trained with CTC.
Both models, while having different
search spaces and different weight functions,
are optimizing the same loss, the marginal log loss.
Empirically, with the same feature encoder and the same optimizer,
there is no significant difference between the two in terms of final performance.
However, each type of model benefits from training jointly
with the other in a multitask training approach.
We hope that drawing the connection between
these models will spawn more research in exploring
different search spaces and loss functions.
In future work, we plan to extend this study of segmental models
to word recognition by exploiting other efficiency and performance
trade-offs.



\bibliography{ref,ref-more}

\begin{thebibliography}{10}
\providecommand{\url}[1]{#1}
\csname url@samestyle\endcsname
\providecommand{\newblock}{\relax}
\providecommand{\bibinfo}[2]{#2}
\providecommand{\BIBentrySTDinterwordspacing}{\spaceskip=0pt\relax}
\providecommand{\BIBentryALTinterwordstretchfactor}{4}
\providecommand{\BIBentryALTinterwordspacing}{\spaceskip=\fontdimen2\font plus
\BIBentryALTinterwordstretchfactor\fontdimen3\font minus
  \fontdimen4\font\relax}
\providecommand{\BIBforeignlanguage}[2]{{%
\expandafter\ifx\csname l@#1\endcsname\relax
\typeout{** WARNING: IEEEtran.bst: No hyphenation pattern has been}%
\typeout{** loaded for the language `#1'. Using the pattern for}%
\typeout{** the default language instead.}%
\else
\language=\csname l@#1\endcsname
\fi
#2}}
\providecommand{\BIBdecl}{\relax}
\BIBdecl

\bibitem{J1976}
F.~Jelinek, ``Continuous speech recognition by statistical methods,''
  \emph{Proceedings of the IEEE}, vol.~64, no.~4, pp. 532--556, 1976.

\bibitem{rabiner1989tutorial}
L.~Rabiner, ``A tutorial on hidden {M}arkov models and selected applications in
  speech recognition,'' \emph{Proceedings of the IEEE}, vol.~77, no.~2, pp.
  257--286, 1989.

\bibitem{bahl1983}
L.~Bahl, F.~Jelinek, and R.~Mercer, ``A maximum likelihood approach to speech
  recognition,'' \emph{IEEE Transactions on Pattern Analysis and Machine
  Intelligence}, vol.~5, pp. 179--190, 1983.

\bibitem{FHJP2013}
E.~Fosler-Lussier, Y.~He, P.~Jyothi, and R.~Prabhavalkar, ``Conditional random
  fields in speech, audio, and language processing,'' \emph{Proceedings of the
  IEEE}, vol. 101, no.~5, pp. 1054--1075, 2013.

\bibitem{smith2001speech}
N.~Smith and M.~Gales, ``{Speech recognition using SVMs},'' in \emph{Advances
  in neural information processing systems (NIPS)}, 2001.

\bibitem{GZ1986}
V.~Zue, J.~Glass, M.~Phillips, and S.~Seneff, ``Acoustic segmentation and
  phonetic classification in the {SUMMIT} system,'' in \emph{IEEE International
  Conference on Acoustics, Speech, and Signal Processing (ICASSP)}, 1989.

\bibitem{de2007template}
M.~De~Wachter, M.~Matton, K.~Demuynck, P.~Wambacq, R.~Cools, and
  D.~Van~Compernolle, ``Template-based continuous speech recognition,''
  \emph{IEEE Transactions on Audio, Speech, and Language Processing}, vol.~15,
  no.~4, pp. 1377--1390, 2007.

\bibitem{BK1985}
M.~A. Bush and G.~E. Kopec, ``Network-based connected digit recognition using
  vector quantization,'' in \emph{IEEE International Conference on Acoustics,
  Speech, and Signal Processing (ICASSP)}, 1985.

\bibitem{zweig2009segmental}
G.~Zweig and P.~Nguyen, ``A segmental {CRF} approach to large vocabulary
  continuous speech recognition,'' in \emph{IEEE Workshop on Automatic Speech
  Recognition and Understanding (ASRU)}, 2009.

\bibitem{CS1997}
G.~Chung and S.~Seneff, ``Hierarchical duration modelling for speech
  recognition using the {ANGIE} framework.'' in \emph{Eurospeech}, 1997.

\bibitem{H2005}
M.~Hasegawa-Johnson, J.~Baker, S.~Borys, K.~Chen, E.~Coogan, S.~Greenberg,
  A.~Juneja, K.~Kirchhoff, K.~Livescu, S.~Mohan \emph{et~al.}, ``Landmark-based
  speech recognition: Report of the 2004 {Johns Hopkins} summer workshop,'' in
  \emph{IEEE International Conference on Acoustics, Speech, and Signal
  Processing (ICASSP)}, 2005.

\bibitem{GLFFP1993}
J.~S. Garofolo, L.~F. Lamel, W.~M. Fisher, J.~G. Fiscus, and D.~S. Pallett,
  ``{DARPA TIMIT} acoustic-phonetic continous speech corpus cd-rom. nist speech
  disc 1-1.1,'' \emph{NASA STI/Recon technical report}, vol.~93, 1993.

\bibitem{glass2003probabilistic}
J.~R. Glass, ``A probabilistic framework for segment-based speech
  recognition,'' \emph{Computer Speech \& Language}, vol.~17, no.~2, pp.
  137--152, 2003.

\bibitem{zweig2012classification}
G.~Zweig, ``Classification and recognition with direct segment models,'' in
  \emph{IEEE International Conference on Acoustics, Speech and Signal
  Processing (ICASSP)}, 2012.

\bibitem{HF2012}
Y.~He and E.~Fosler-Lussier, ``Efficient segmental conditional random fields
  for phone recognition,'' in \emph{INTERSPEECH}, 2012.

\bibitem{abdel2013deep}
O.~Abdel-Hamid, L.~Deng, D.~Yu, and H.~Jiang, ``Deep segmental neural networks
  for speech recognition,'' in \emph{INTERSPEECH}, 2013.

\bibitem{tang2015discriminative}
H.~Tang, W.~Wang, K.~Gimpel, and K.~Livescu, ``Discriminative segmental
  cascades for feature-rich phone recognition,'' in \emph{IEEE Workshop on
  Automatic Speech Recognition and Understanding (ASRU)}, 2015.

\bibitem{lu2016segmental}
L.~Lu, L.~Kong, C.~Dyer, N.~A. Smith, and S.~Renals, ``Segmental recurrent
  neural networks for end-to-end speech recognition,'' in \emph{INTERSPEECH},
  2016.

\bibitem{ostendorf1996}
M.~Ostendorf, V.~Digalakis, and O.~Kimball, ``From {HMM}'s to segment models: A
  unified view of stochastic modeling for speech recognition,'' \emph{IEEE
  Transactions on Speech and Audio Processing}, pp. 360--378, 1996.

\bibitem{sarawagi2004semi}
S.~Sarawagi and W.~W. Cohen, ``Semi-{Markov} conditional random fields for
  information extraction.'' in \emph{Advances in Neural Information Processing
  Systems (NIPS)}, vol.~17, 2004.

\bibitem{ZG2013}
S.-X. Zhang and M.~J. Gales, ``Structured {SVM}s for automatic speech
  recognition,'' \emph{IEEE Transactions on Audio, Speech, and Language
  Processing}, vol.~21, no.~3, pp. 544--555, 2013.

\bibitem{kong2016segmental}
L.~Kong, C.~Dyer, and N.~A. Smith, ``Segmental recurrent neural networks,'' in
  \emph{International Conference on Learning Representations (ICLR)}, 2016.

\bibitem{hochreiter1997long}
S.~Hochreiter and J.~Schmidhuber, ``Long short-term memory,'' \emph{Neural
  computation}, vol.~9, no.~8, pp. 1735--1780, 1997.

\bibitem{M1997}
M.~Mohri, ``Finite-state transducers in language and speech processing,''
  \emph{Computational linguistics}, vol.~23, no.~2, pp. 269--311, 1997.

\bibitem{M2002}
------, ``Semiring frameworks and algorithms for shortest-distance problems,''
  \emph{Journal of Automata, Languages and Combinatorics}, vol.~7, no.~3, pp.
  321--350, 2002.

\bibitem{tang2016end}
H.~Tang, W.~Wang, K.~Gimpel, and K.~Livescu, ``End-to-end training approaches
  for discriminative segmental models,'' in \emph{IEEE Workshop on Spoken
  Language Technology (SLT)}, 2016.

\bibitem{TGL2014}
H.~Tang, K.~Gimpel, and K.~Livescu, ``A comparison of training approaches for
  discriminative segmental models,'' in \emph{INTERSPEECH}, 2014.

\bibitem{HDSN2008}
G.~Heigold, T.~Deselaers, R.~Schl\"uter, and H.~Ney, ``Modified {MMI}/{MPE}: A
  direct evaluation of the margin in speech recognition,'' in
  \emph{International Conference on Machine learning (ICML)}, 2008.

\bibitem{MWN2010}
E.~McDermott, S.~Watanabe, and A.~Nakamura, ``Discriminative training based on
  an integrated view of {MPE} and {MMI} in margin and error space,'' in
  \emph{IEEE International Conference on Acoustics Speech and Signal Processing
  (ICASSP)}, 2010.

\bibitem{lu2017multi}
L.~Lu, L.~Kong, C.~Dyer, and N.~A. Smith, ``Multi-task learning with {CTC} and
  segmental {CRF} for speech recognition,'' \emph{CoRR}, vol. abs/1702.06378,
  2017.

\bibitem{graves2006connectionist}
A.~Graves, S.~Fern{\'a}ndez, F.~Gomez, and J.~Schmidhuber, ``Connectionist
  temporal classification: labelling unsegmented sequence data with recurrent
  neural networks,'' in \emph{International Conference on Machine Learning
  (ICML)}, 2006.

\bibitem{P2011}
D.~Povey, A.~Ghoshal, G.~Boulianne, L.~Burget, O.~Glembek, N.~Goel,
  M.~Hannemann, P.~Motlicek, Y.~Qian, P.~Schwarz \emph{et~al.}, ``The {Kaldi}
  speech recognition toolkit,'' in \emph{IEEE Workshop on Automatic Speech
  Recognition and Understanding (ASRU)}, 2011.

\bibitem{L1988}
K.-F. Lee, ``On large-vocabulary speaker-independent continuous speech
  recognition,'' \emph{Speech communication}, vol.~7, no.~4, pp. 375--379,
  1988.

\bibitem{VDH2013}
V.~Vanhoucke, M.~Devin, and G.~Heigold, ``Multiframe deep neural networks for
  acoustic modeling,'' in \emph{IEEE International Conference on Acoustics,
  Speech and Signal Processing (ICASSP)}, 2013.

\bibitem{MLWZG2015}
Y.~Miao, J.~Li, Y.~Wang, S.~Zhang, and Y.~Gong, ``Simplifying long short-term
  memory acoustic models for fast training and decoding,'' in \emph{IEEE
  International Conference on Acoustics, Speech and Signal Processing
  (ICASSP)}, 2015.

\bibitem{ZSV2014}
W.~Zaremba, I.~Sutskever, and O.~Vinyals, ``Recurrent neural network
  regularization,'' \emph{CoRR}, vol. abs/1409.2329, 2014.

\bibitem{GB2010}
X.~Glorot and Y.~Bengio, ``Understanding the difficulty of training deep
  feedforward neural networks,'' in \emph{AISTATS}, 2010.

\bibitem{TH2012}
T.~Tieleman and G.~Hinton, ``Lecture 6.5-{RMS}prop: Divide the gradient by a
  running average of its recent magnitude,'' \emph{COURSERA: Neural networks
  for machine learning}, 2012.

\bibitem{TWGL2016}
H.~Tang, W.~Wang, K.~Gimpel, and K.~Livescu, ``Efficient segmental cascades for
  speech recognition,'' in \emph{INTERSPEECH}, 2016.

\bibitem{DHSN2016}
P.~Doetsch, S.~Hegselmann, R.~Schl\"uter, and H.~Ney, ``Inverted {HMM} -- a
  proof of concept,'' in \emph{NIPS 2016 End-to-end Learning for Speech and
  Audio Processing Workshop}, 2016.

\bibitem{H2015}
Y.~He, ``Segmental models with an exploration of acoustic and lexical grouping
  in automatic speech recognition,'' Ph.D. dissertation, The Ohio State
  University, 2015.

\bibitem{T2015}
L.~T{\'o}th, ``Phone recognition with hierarchical convolutional deep maxout
  networks,'' \emph{EURASIP Journal on Audio, Speech, and Music Processing},
  vol. 2015, no.~1, p.~25, 2015.

\bibitem{HG1998}
A.~Halberstadt and J.~Glass, ``Heterogeneous measurements and multiple
  classifiers for speech recognition,'' in \emph{International Conference on
  Spoken Language Processing}, 1998.

\bibitem{CPS2016}
R.~Collobert, C.~Puhrsch, and G.~Synnaeve, ``Wav2letter: an end-to-end
  convnet-based speech recognition system,'' \emph{CoRR}, vol. abs/1609.03193,
  2016.

\bibitem{CBSCB2015}
J.~K. Chorowski, D.~Bahdanau, D.~Serdyuk, K.~Cho, and Y.~Bengio,
  ``Attention-based models for speech recognition,'' in \emph{Advances in
  Neural Information Processing Systems (NIPS)}, 2015.

\bibitem{BCSBB2016}
D.~Bahdanau, J.~Chorowski, D.~Serdyuk, P.~Brakel, and Y.~Bengio, ``End-to-end
  attention-based large vocabulary speech recognition,'' in \emph{IEEE
  International Conference on Acoustics, Speech and Signal Processing
  (ICASSP)}, 2016.

\bibitem{CJLV2016}
W.~Chan, N.~Jaitly, Q.~Le, and O.~Vinyals, ``Listen, attend and spell: A neural
  network for large vocabulary conversational speech recognition,'' in
  \emph{IEEE International Conference on Acoustics, Speech and Signal
  Processing (ICASSP)}, 2016.

\bibitem{povey2016purely}
D.~Povey, V.~Peddinti, D.~Galvez, P.~Ghahrmani, V.~Manohar, X.~Na, Y.~Wang, and
  S.~Khudanpur, ``{Purely sequence-trained neural networks for {ASR} based on
  lattice-free {MMI}},'' in \emph{INTERSPEECH}, 2016.

\bibitem{G2012}
A.~Graves, ``Sequence transduction with recurrent neural networks,''
  \emph{CoRR}, vol. abs/1211.3711, 2012.

\bibitem{he2015segmental}
Y.~He and E.~Fosler-Lussier, ``Segmental conditional random fields with deep
  neural networks as acoustic models for first-pass word recognition.'' in
  \emph{INTERSPEECH}, 2015.

\bibitem{graves2014towards}
A.~Graves and N.~Jaitly, ``Towards end-to-end speech recognition with recurrent
  neural networks,'' in \emph{International Conference on Machine Learning
  (ICML)}, 2014.

\bibitem{MXJN2015}
A.~L. Maas, Z.~Xie, D.~Jurafsky, and A.~Y. Ng, ``Lexicon-free conversational
  speech recognition with neural networks,'' in \emph{Human Language
  Technologies: The Annual Conference of the North American Chapter of the
  ACL}, 2015.

\bibitem{miao2015eesen}
Y.~Miao, M.~Gowayyed, and F.~Metze, ``{EESEN}: End-to-end speech recognition
  using deep {RNN} models and {WFST}-based decoding,'' in \emph{IEEE Workshop
  on Automatic Speech Recognition and Understanding (ASRU)}, 2015.

\bibitem{MMONN2012}
A.~L. Maas, S.~D. Miller, T.~M. O'neil, A.~Y. Ng, and P.~Nguyen, ``Word-level
  acoustic modeling with convolutional vector regression,'' in \emph{ICML
  Workshop on Representation Learning}, 2012.

\bibitem{BH2014}
S.~Bengio and G.~Heigold, ``Word embeddings for speech recognition,'' in
  \emph{INTERSPEECH}, 2014.

\bibitem{SLS2016}
H.~Soltau, H.~Liao, and H.~Sak, ``Neural speech recognizer: Acoustic-to-word
  {LSTM} model for large vocabulary speech recognition,'' \emph{CoRR}, vol.
  abs/1610.09975, 2016.

\bibitem{ARSPN2017}
K.~Audhkhasi, B.~Ramabhadran, G.~Saon, M.~Picheny, and D.~Nahamoo, ``Direct
  acoustics-to-word models for english conversational speech recognition,''
  \emph{CoRR}, vol. abs/1703.07754, 2017.

\end{thebibliography}
\bibliographystyle{IEEEtran}

%
%
%
%
%
%
%

\end{document}